\pdfoutput=1

\documentclass[11pt]{article}

\usepackage[final]{acl}

\usepackage{times}
\usepackage{latexsym}
\usepackage{amssymb}

\usepackage[T1]{fontenc}

\usepackage[utf8]{inputenc}

\usepackage{microtype}

\usepackage{inconsolata}

\usepackage{graphicx}

\usepackage{enumitem}
\usepackage{longtable}
\usepackage{csquotes}
\usepackage[a4paper]{geometry}
\usepackage{booktabs}
\usepackage{amsmath}

\usepackage{listings}
\usepackage{xcolor}

\lstdefinelanguage{json}{
    basicstyle=\ttfamily\small,
    numbers=none,
    numberstyle=\tiny\color{gray},
    stepnumber=1,
    numbersep=5pt,
    showstringspaces=false,
    breaklines=true,
    frame=single,
    backgroundcolor=\color{black!5},
    keywordstyle=\color{blue},
    stringstyle=\color{red},
    morestring=[b]",
    morecomment=[l]{//},
    commentstyle=\color{gray},
    morekeywords={"entity_groups", "name", "segment", "tabulate", "header", "entities", "probability", "type", "fontVariance", "color"} 
}

\usepackage{fancyhdr}
\fancypagestyle{firstpage}{
  \fancyhf{}                    
  \fancyhead[L]{\normalsize Accepted at EMNLP 2025} 
  
}

\title{FlexDoc: Parameterized Sampling for Diverse Multilingual Synthetic Documents for Training Document Understanding Models}

\author{
\textbf{Karan Dua\textsuperscript{}},
\textbf{Hitesh Laxmichand Patel\textsuperscript{}},
\textbf{Puneet Mittal\textsuperscript{}},
\textbf{Ranjeet Gupta\textsuperscript{}},
\textbf{Amit Agarwal\textsuperscript{}}, \\
\textbf{Praneet Pabolu\thanks{Work done during employment with Oracle Corporation}}, 
\textbf{Srikant Panda\footnotemark[1]},
\textbf{Hansa Meghwani\textsuperscript{}},
\textbf{Graham Horwood\textsuperscript{}},
\textbf{Fahad Shah\textsuperscript{}}
\\ 
\\
 \textsuperscript{}Oracle AI
\\
 \small{
   \textbf{Correspondence:} \href{karan.dua@oracle.com}{karan.dua@oracle.com}
 }
}

\begin{document}
\maketitle
\thispagestyle{firstpage}
\begin{abstract}
Developing document understanding models at enterprise scale requires large, diverse, and well-annotated datasets spanning a wide range of document types. However, collecting such data is prohibitively expensive due to privacy constraints, legal restrictions, and the sheer volume of manual annotation needed - costs that can scale into millions of dollars. We introduce \textbf{FlexDoc}, a scalable synthetic data generation framework that combines \textbf{Stochastic Schemas} and \textbf{Parameterized Sampling} to produce realistic, multilingual semi-structured documents with rich annotations. By probabilistically modeling layout patterns, visual structure, and content variability, FlexDoc enables the controlled generation of diverse document variants at scale. Experiments on Key Information Extraction (KIE) tasks demonstrate that FlexDoc-generated data improves the absolute F1 Score by up to \textbf{11\%} when used to augment real datasets, while reducing annotation effort by over \textbf{90\%} compared to traditional hard-template methods. The solution is in active deployment, where it has accelerated the development of enterprise-grade document understanding models while significantly reducing data acquisition and annotation costs.
\end{abstract}

\section{Introduction}
\subsection{Document Understanding}
Document Understanding refers to the task of automatically interpreting and extracting structured information from documents that combine text, layout, and visual elements \cite{agarwal-etal-2025-fs,agarwal2023pseudo}. Unlike plain-text NLP, it requires reasoning over both content and structure, making it inherently multimodal. In enterprise and industrial settings, this capability enables critical workflows such as invoice processing, claims automation, identity verification, and compliance auditing. Common applications include key information extraction, document classification, and form or table understanding - tasks that are especially relevant in data-intensive domains like finance, healthcare, insurance, and business process automation \cite{pattnayak2024survey,pattnayak2025clinicalqa20multitask,panda2025out,panda2025techniques}.

\subsection{Multi-Modal Document Understanding Models}

The spatial position and context of elements are critical for document understanding. Recent multimodal models \cite{Xu_2020, yang, anoop, huang2022layoutlmv3pretrainingdocumentai} combine textual, visual, and positional features to capture both content and layout. Vision-language models like Phi-4-Multimodal \cite{microsoft2024phi4multi}, Qwen-VL \cite{Qwen-VL} and LLaVA \cite{liu2023visualinstructiontuning} further improve spatial reasoning through techniques like M-RoPE and image-text alignment.

Training these models, however, requires large, diverse annotated datasets - often impractical to collect due to privacy, legal, and cost constraints. Synthetic data generation offers a scalable alternative, enabling controlled generation without the limitations of real-world data acquisition.

\begin{figure*}[h!]
  \includegraphics[width=1.0\linewidth]{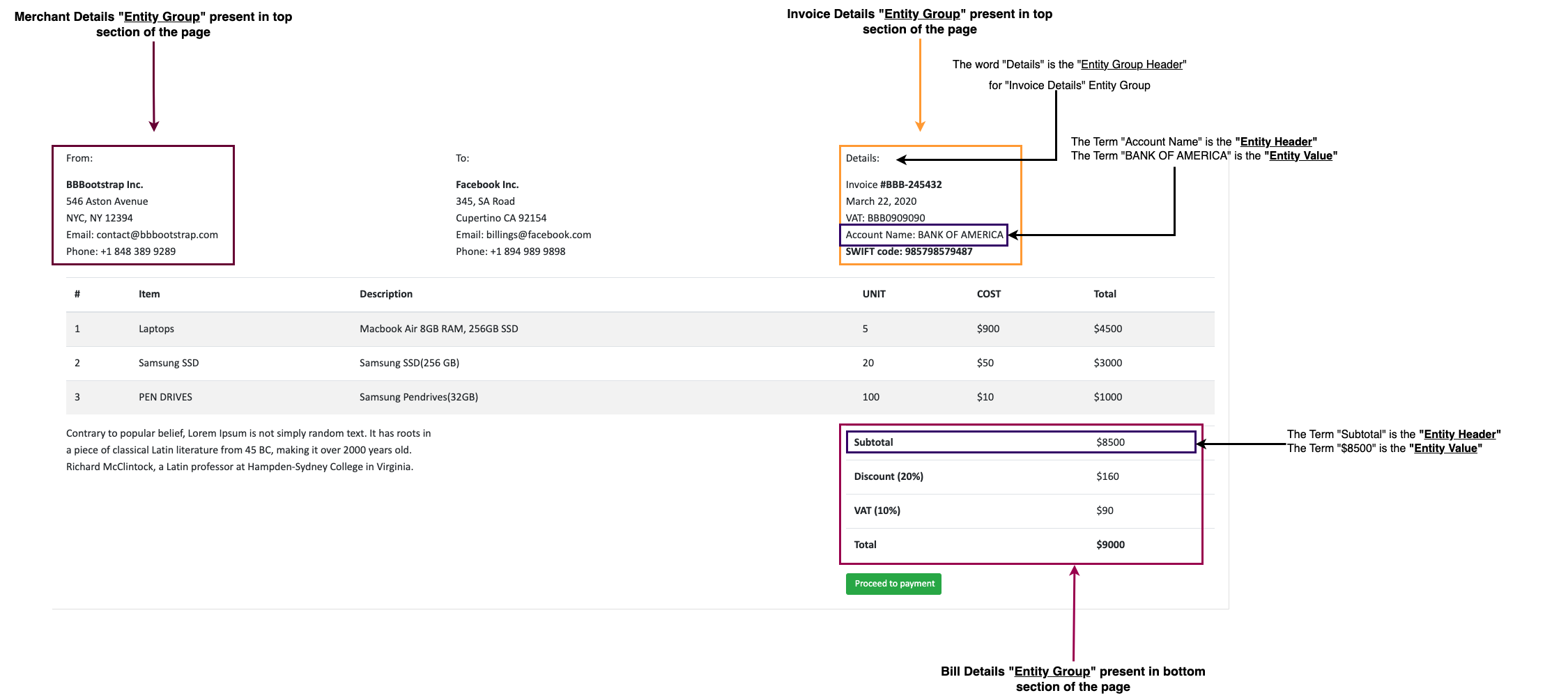}
  \caption {\label{fig:invoice-patterns}Typical patterns in an invoice}
\end{figure*}

\subsection{Hard Template Based Document Generators}
A common approach to synthetic document generation uses hard template-based methods \cite{monsur-etal-2023-synthnid, capobianco2017docemultoolkitgeneratestructured, DBLP:journals/corr/abs-2108-02899,agarwal2024synthetic,agarwal2024techniques}. This typically involves collecting real documents, manually annotating and whiting out sensitive values (e.g., names, addresses), and replacing them with type-consistent fake values, as illustrated in Figure~\ref{fig:fully-templatic} in appendix. While effective for highly structured documents such as IDs and passports \cite{monsur-etal-2023-synthnid, cui2021document, jaume2019, 10.1145/3474085.3475345, 7801919, Xu2020LayoutXLMMP, zhang-etal-2021-entity, cvpr2023geolayoutlm, lee-etal-2022-formnet,patel2024llm,agarwal2024enhancing}, this method has notable limitations when applied to semi-structured documents. The layout diversity is constrained by the original templates, limiting positional variation of fields like merchant names or table sizes \cite{teakgyu-et-al}. Predefined whiteout regions often fail to accommodate longer replacement values, and inserting new content can degrade visual fidelity. Moreover, the approach is difficult to scale, as each template requires collecting and manually annotating real documents. Figure~\ref{fig:fully-templatic} in appendix describes the hard template based approach with an example.

\subsection{Privacy}
Data collection and its use is subject to several privacy challenges, including but not limited to data use restrictions, data regulations such as GDPR and ethical considerations to ensure diverse representations while avoiding bias. Models trained on real-world data may also inadvertently reveal private information during inference.
\vspace{4pt}

To address the above challenges, we introduce \textbf{FlexDoc} - a framework to generate diverse, multilingual, annotated synthetic document datasets for training Document Understanding models. Specifically:

\begin{itemize}[itemsep=0pt, topsep=2pt]
    \item We introduce a novel algorithm called \textbf{Parameterized Sampling} centered around \textbf{Stochastic Schemas} which can generate hundreds of thousands of unique semi-structured documents using a single definition, \textbf{along with annotations} (key value labels, bounding boxes, document types, table boundaries etc.) with \textbf{guaranteed accuracy}.
    \item We also present a novel \textbf{Dynamic Virtual Grid Algorithm} that organizes document elements into non-overlapping regions while enhancing visual diversity.
    \item The generation process uses a fake value generator, eliminating privacy risks and can be configured for locale-specific tuning.
    \item The pipeline is \textbf{multilingual}. The same stochastic schema written in English can be used to generate documents in different languages by switching just two simple configurations.

\end{itemize}

\begin{figure*}[h!]
  \includegraphics[width=1.0\linewidth]{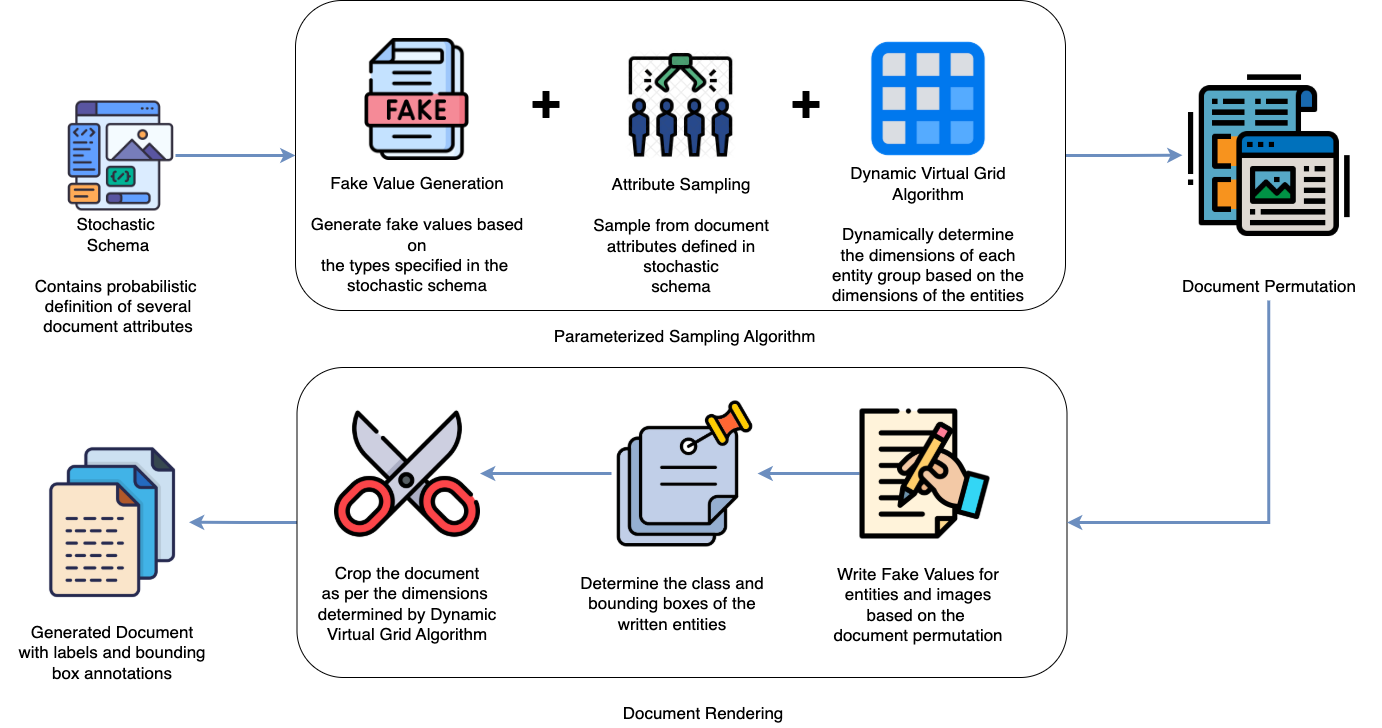}
  \caption {\label{fig:pipeline-highlevel}High-Level description of FlexDoc for generating Synthetic Annotated Documents}
\end{figure*}

\section{Related Work}

Recent work has explored using large language models (LLMs) for synthetic data generation. \citet{dua2025speechweave} introduced an end to end pipeline for generating synthetic data for training speech models using LLMs and speech audio generation and voice standardization models. \citet{josifoski2023synthie} introduced SynthIE, prompting LLMs to generate input-output pairs for information extraction (IE) without labeled data. GuideX \citep{dela2025guidex} generates schema-guided examples for fine-tuning LLaMA 3.1 \cite{pattnayak2025tokenizationmattersimprovingzeroshot,agarwal2025aligningllmsmultilingualconsistency}, while \citet{bhattacharyya2024task_aware_labeling} distill soft labels from multimodal models like Claude 3 into compact KIE models. In specialized domains, \citet{woo2024synthetic_clinical_distill} synthesize clinical Q\&A pairs, showing that distilled models can rival their teachers. 

For visually rich documents, most approaches rely on hard templates. SynthNID \citep{monsur-etal-2023-synthnid} overlays fake values onto ID templates, while Genalog \citep{DBLP:journals/corr/abs-2108-02899} uses HTML/CSS templates with synthetic content and degradation steps. \citet{RAMAN2022108660} explored variational templates by treating document components as random variables, though their scope was limited to layout recognition.

On the modeling side, LayoutLM \citep{Xu_2020, huang2022layoutlmv3pretrainingdocumentai} pioneered multimodal pretraining with text and layout features. More recent models like Qwen-VL 2.5 \citep{bai2023qwenvl}, LLaMA 3.2 Vision, Phi-4 Multimodal \citep{microsoft2024phi4multi}, and LLaVA \citep{liu2023llava} advance spatial reasoning through visual-text alignment and hard negative mining \cite{meghwani-etal-2025-hard}. GPT-4o \cite{openai2024gpt4ocard, chen2025empiricalstudygpt4oimage, yan2025gptimgevalcomprehensivebenchmarkdiagnosing} introduces native image generation with improved attribute binding and text clarity, though it lacks fine-grained control and diversity beyond prompt variations.

\section{Methodology}
The subsequent sections detail the rationale behind FlexDoc and systematically explain its components. The detailed algorithm is described in Figure~\ref{fig:overall-algorithm} in appendix.
\subsection{Intuition}
\subsubsection{Patterns in Semi-Structured Documents}
The format of information within semi-structured documents like invoices and receipts often adheres to identifiable patterns. Typically, the data within such documents tends to display the following characteristics:
\begin{enumerate}[itemsep=0pt, topsep=2pt]
    \item Documents contain known identifiable elements (e.g., \textbf{Merchant Name}, \textbf{Invoice Date}) with specific types (text, date, number). We refer to each as an \textbf{entity}, and its value as the \textbf{entity value}.
    
    \item Entities typically follow specific header texts (e.g., \textbf{Merchant Name}, \textbf{Sold by}), which guide accurate identification. We call this the \textbf{entity header}.
    
    \item Entities are grouped (e.g., Merchant Details, Invoice Details), forming \textbf{entity groups}.
    
    \item These groups may also be introduced by header text (e.g., \textbf{Seller Information}), called \textbf{entity group headers}.
    
    \item Documents exhibit structured layouts with consistent fonts, colors, and alignments across entities and groups.
\end{enumerate}
Figure~\ref{fig:invoice-patterns} describes such properties with an example invoice.
\subsubsection{Variations in Semi-Structured Documents}
While semi-structured documents often follow recognizable patterns, they also exhibit significant variation in entities, entity groups, headers, and layout structures - adding complexity to document understanding. Based on our analysis across diverse document types, we observe the following variations:

\begin{enumerate}[itemsep=0pt, topsep=2pt]
    \item Entity groups can appear anywhere on the page (e.g., Merchant Details may be at the top or bottom).
    \item Placement within a section is not fixed (e.g., Customer Details may appear top-left, top-right, or center).
    
    \item Some documents may omit certain entity groups (e.g., Shipping Details or Payment Terms may be absent).
    
    \item Not all entity groups include headers (e.g., Merchant Details may lack a header).
    
    \item Headers, if present, vary in wording (e.g., Customer Details may appear as \textbf{Buyer Details} or \textbf{Client Details}).
    
    \item Entity groups can be formatted differently - stacked, tabular (vertical/horizontal), or mixed - with varied table structures and formatting.
    
    \item Some entities may be missing (e.g., Invoice Due Date or Purchase Order Number).
    
    \item Entities may or may not have headers (e.g., Customer Name might lack a label).
    
    \item Headers for entities vary in wording (e.g., \textbf{Subtotal} vs. \textbf{Sub-Total Amount}).
    
    \item The order of entities within a group is inconsistent (e.g., email and phone order varies in Customer Details).
    
    \item Fonts and colors differ across documents and between entity headers, group headers, and values.
    
    \item Entity values differ widely (e.g., different customer names).
    
    \item Entities may be left, right, or center-aligned based on context.
\end{enumerate}

Figure~\ref{fig:doc-variations} in appendix depicts these variations by comparing two invoices side by side.

\subsection{Framework/Algorithm}

The FlexDoc algorithm, described in Figure~\ref{fig:pipeline-highlevel}, consists of three major components: A \textbf{Stochastic Schema}, a \textbf{Parameterized Sampling} algorithm, and a \textbf{Document Rendering} algorithm. 

\subsubsection{Stochastic Schema}

Building on identified patterns and variations, we construct stochastic schemas - where element properties are defined as random variables, with either specified distributions or value ranges. For example, rather than fixing the number of rows in the Item Details table, we use a uniform distribution and sample a value at generation time.

Stochastic schemas can also define the presence of elements probabilistically. For instance, the Customer Phone Number entity may appear in a document based on a predefined probability, reflecting real-world variability. Additionally, stochastic schemas specify style and structural attributes - like fonts, colors, and grid layouts - under a shared configuration.

A document can be defined by a stochastic schema \( \mathcal{T}_s \), consisting of entity groups \( \mathcal{G} = \{G_1, \dots, G_K\} \). Each entity group \( G_k \) is defined as:
\[
G_k = (\mathcal{E}_k, \mathcal{H}_k, p_k, \alpha_k)
\]
\begin{itemize}
  \setlength\itemsep{0pt}
  \setlength\topsep{0pt}
  \item \( \mathcal{E}_k = \{e_{k1}, \dots, e_{kn}\} \): set of entities
  \item \( \mathcal{H}_k \): set of possible group headers
  \item \( p_k \in [0,1] \): probability of group appearance
  \item \( \alpha_k \): layout attributes (e.g., preferred sections, alignment, table format)
\end{itemize}

Each entity \( e_{ki} \in \mathcal{E}_k \) is a tuple:
\[
e_{ki} = (T_{ki}, \mathcal{H}_{ki}, p_{ki}, \beta_{ki})
\]
\begin{itemize}
  \setlength\itemsep{0pt}
  \setlength\topsep{0pt}
  \item \( T_{ki} \): entity type (e.g., name, address)
  \item \( \mathcal{H}_{ki} \): header label variants
  \item \( p_{ki} \in [0,1] \): probability of entity appearance
  \item \( \beta_{ki} \): visual/layout attributes
\end{itemize}

These schemas, defined as JSONs, serve as loose specifications. At runtime, the algorithm samples from various distributions to generate diverse document permutations, enabling the creation of hundreds of thousands of unique documents. Appendix~\ref{sec:appendix-template-definition} provides an example of an entity group definition and describes related attributes and style configuration.

\begin{figure*}[h!]
\includegraphics[width=1.0\linewidth]{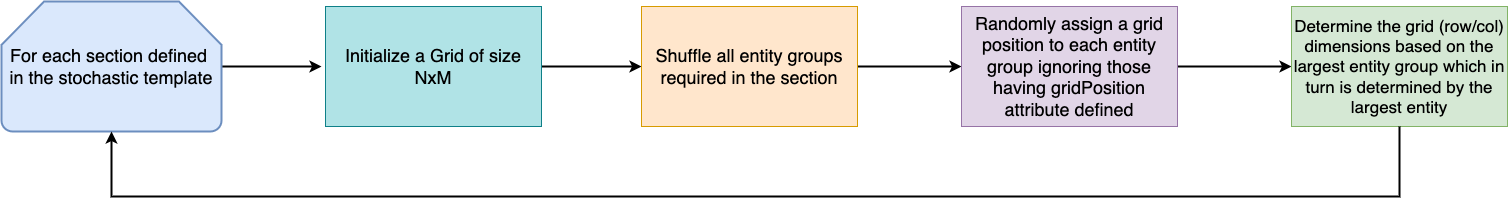}
\caption {\label{fig:virtual-grid-algo}Dynamic Virtual Grid Algorithm}
\end{figure*}

\subsubsection{Parameterized Sampling Algorithm}

\paragraph{Entity Fake Value Generation:}

Entity values are generated using a type-specific value generator:
\[
v_{ki} = f_{T_{ki}}(\theta)
\]
where \( f_{T_{ki}} \) denotes a generator function selected based on the entity type \( T_{ki} \), and \( \theta \) includes generation parameters such as locale, format, and value constraints.

To implement this, we use the Python library Faker~\cite{joke2kfaker}, along with custom fake value generators. Faker supports generating synthetic data for various entities (e.g., names, addresses, phone numbers) across multiple locales.

Each stochastic schema includes a generator class defined within the same JSON schema. A typical generator class definition is shown below:

\begin{verbatim}
"fake_value_generator_class": 
"utils.doc_generator.InvoiceGenerator"
\end{verbatim}

The type of each entity defined in stochastic schema is utilized by the generator class to produce entity values. For e.g. a merchant name may be of type \textbf{name}, while a merchant address may be of type \textbf{address}.

\paragraph{Attribute Sampling:}
We utilize a sampling algorithm to freeze the stochastic definitions defined in the schema. Each instance of sampling these attributes generates a \textbf{document permutation}. A document permutation is a frozen outline of all the stochastic attributes required in a document. From a single stochastic schema, thousands of distinct document permutations can be generated. An instance of a document permutation is created by simply replacing fake values for each entity.

A document permutation \( \mathcal{T}_p \) is generated by freezing the stochastic schema \( \mathcal{S} \):
\[
\mathcal{T}_p  = \left\{
G_{k}, e_{ki} | G_{k} , v_{ki}, \alpha_k, \beta_{ki}, H_k, H_{ki}, \ldots
\right\}
\]

\textit{where each component - entity/group presence, value, layout, and style - is sampled from schema-defined probabilities or distributions.}
\\

Table~\ref{tab:attribute-sampling} in appendix describes the techniques used to sample the attributes in stochastic schema.

\paragraph{Dynamic Virtual Grid Algorithm:}

Documents often consist of multiple sections, each of which may contain numerous entity groups. While attribute sampling determines the designated section for each entity group, placing those groups within the sections and writing on the canvas poses several challenges: random positioning can cause overlaps, fixed sizes limit layout flexibility, and sequential placement may disrupt alignment.

To address this, we introduce a \textbf{Dynamic Virtual Grid Arrangement} algorithm. Each section is treated as a virtual grid whose dimensions are schema-driven. Entity groups are placed into cells, and row/column sizes are dynamically adjusted to minimize whitespace and preserve visual structure.

Figure~\ref{fig:virtual-grid-algo} provides an overview; a detailed example is described in Figure~\ref{fig:virtual-grid-algo-example} in appendix.

\subsubsection{Document Rendering Algorithm}
Let the document canvas be divided into a grid \( \mathcal{C} \in \mathbb{R}^{H \times W} \), with each entity group assigned to a grid cell via the mapping:
\[
\phi: G_k \rightarrow (i, j)
\]
where \( G_k \) is the \(k\)-th entity group and \( (i, j) \) its target cell on the canvas.

We define a rendering function \( \mathcal{R}(\mathcal{T}_h, \phi) \) that draws sampled entity groups onto a blank canvas based on grid positions and layout attributes from the stochastic schema \( \mathcal{T}_h \), while recording bounding boxes for annotation.

Rendering uses the Pillow library~\cite{clark2015pillow}, respecting the Dynamic Virtual Grid Arrangement and frozen layout constraints to ensure consistent structure and accurate annotations. A high-level overview is provided in Figure~\ref{fig:doc-rendering} in appendix.

\vspace{5pt}

Final document and annotation output:
\[
\mathcal{I} = \mathcal{R}(\mathcal{T}_h, \phi), \quad \text{Ann} = \left\{ (b_{ki}, v_{ki}), (\text{class labels}) \right\}
\]

Here, \( \mathcal{I} \) is the rendered image, and \( \text{Ann} \) the set of annotations - each containing a bounding box \( b_{ki} \), value \( v_{ki} \), and class labels, which vary by task (e.g., per-entity labels in KIE).

\subsection{Multilinguality}
The algorithm can be configured to generate data in different languages (one language per document) while using the same stochastic schema defined in English. Configuration \textit{faker\_locale} switches the faker locale to generate entity values in the target language while \texttt{\{"translation": \{"enable": "True", "target\_lang\_code": "\textless lang\_code\textgreater"\}\}} uses a machine translation engine to translate entity and entity group headers defined in the JSON schema to the target language. 

Example documents along with KIE specific annotations generated using FlexDoc are available in Appendices~\ref{sec:appendix-annotation} and \ref{sec:generated-samples}.

\section{Evaluation Results}
\label{sec:evaluation}
Experimentation settings, including the choice of models and datasets for evaluation are thoroughly detailed in Appendix~\ref{sec:experimentation-settings}.
\subsection{Downstream Model Performance}
\begin{table}[ht]
\footnotesize
\renewcommand{\arraystretch}{1.2}
\centering
\begin{tabular}{p{3.3cm}cc}
  \hline
  \textbf{Train Dataset} & \textbf{LayoutLM} & \textbf{Phi-4} \\
  \hline
  Zero Shot & NA & 31.2 \\
  Synthetic 5k Only & 54.1 & 53.3 \\
  DocILE Only (Baseline) & 72.7 & 74.6 \\
  DocILE + 1k Synthetic & 76.3 & 79.8 \\
  DocILE + 2k Synthetic & 78.8 & 82.4 \\
  DocILE + 3k Synthetic & 81.8 & 83.1 \\
  DocILE + 4k Synthetic & 82.7 & 85.3 \\
  DocILE + 5k Synthetic & \textbf{82.9} & \textbf{85.6} \\
  \hline
\end{tabular}
\caption{\label{tab:docile-results}
  F1 scores when incrementally adding FlexDoc-generated synthetic data to the DocILE dataset.
}
\end{table}
\begin{table}[ht]
\footnotesize
\renewcommand{\arraystretch}{1.2}
\centering
\begin{tabular}{p{3.3cm}cc}
  \hline
  \textbf{Train Dataset} & \textbf{LayoutLM} & \textbf{Phi-4} \\
  \hline
  Zero Shot & NA & 23.1 \\
  Synthetic 5k Only & 47.3 & 44.4 \\
  IDSEM Only (Baseline) & 65.5 & 71.2 \\
  IDSEM + 1k Synthetic & 67.3 & 74.5 \\
  IDSEM + 2k Synthetic & 69.1 & 76.0 \\
  IDSEM + 3k Synthetic & 73.3 & 77.3 \\
  IDSEM + 4k Synthetic & 74.2 & 78.6 \\
  IDSEM + 5k Synthetic & \textbf{75.6} & \textbf{79.5} \\
  \hline
\end{tabular}
\caption{\label{tab:idsem-results}
  F1 scores when incrementally adding FlexDoc-generated synthetic data to the IDSEM dataset (Spanish invoices).
}
\end{table}

\vspace{-8pt}
We assess FlexDoc's effectiveness on the Key Information Extraction (KIE) task - one of the most complex tasks in Document Understanding - using both the DocILE \cite{simsa2023docile} (English) and IDSEM \cite{sanchez2022idsem} (Spanish invoices) datasets. We first train LayoutLM (an encoder-based multimodal model) and Phi-4-Multimodal-Instruct (a decoder-based generative multimodal model) on these real datasets (as baselines), and then incrementally augment them with 1k–5k FlexDoc-generated synthetic samples.

For DocILE, augmenting with synthetic data results in significant performance improvements, with LayoutLM achieving an F1 score of \textbf{82.9\%} and Phi-4 reaching \textbf{85.6\%} after adding 5k synthetic samples - an increase of up to \textbf{10.2\%} for LayoutLM and \textbf{11\%} for Phi-4 against the baseline.

Similarly, on the IDSEM dataset, incrementally adding synthetic data boosts performance significantly, with final F1 scores of \textbf{75.6\%} for LayoutLM and \textbf{79.5\%} for Phi-4 after adding 5k synthetic samples - showing an improvement of \textbf{10.1\%} for LayoutLM and \textbf{8.3\%} for Phi-4 against the baseline.

These results demonstrate FlexDoc's effectiveness in enhancing KIE task performance in both English and Spanish invoice datasets, highlighting its potential for multilingual applications in document understanding tasks.

\subsection{Ablation Study - Dynamic Virtual Grid Algorithm}
We also evaluate the effectiveness of the Dynamic Virtual Grid Algorithm by replacing it with a baseline method that randomly places entities in non-overlapping positions without considering layout structure. When this algorithm is disabled, LayoutLM’s performance (with 5k synthetic documents) drops from \textbf{82.9\%} to \textbf{75.4\%}, and Phi-4's from \textbf{85.9\%} to \textbf{80.6\%}. This highlights the critical role of structured placement in enhancing the quality and utility of synthetic documents for key information extraction tasks.
\subsection{Comparison with Hard-Template Approach}
\vspace{-8pt}
\begin{table}[ht]
\footnotesize
\renewcommand{\arraystretch}{1.1}
\centering
\begin{tabular}{l@{\hskip 6pt}c@{\hskip 6pt}c@{\hskip 6pt}c}
\hline
\textbf{Approach} & \textbf{LayoutLM} & \textbf{Phi-4} & \textbf{Effort} \\
                  & \textbf{(F1)}     & \textbf{(F1)}  & \textbf{(mins)} \\
\hline
Hard Template & \textbf{69.8} & \textbf{72.2} & 1500 \\
FlexDoc       & 67.4          & 69.8          & \textbf{150} \\
\hline
\end{tabular}
\caption{\label{tab:template-flexdoc-results} Performance and annotation effort comparison for Hard Template vs. FlexDoc approaches on Insurance Cards Dataset.}
\end{table}

\vspace{-8pt}
We compare FlexDoc against the widely used hard-template-based generation approach. To assess robustness, we evaluate on a proprietary insurance card dataset, where FlexDoc achieves comparable performance while reducing annotation effort by approximately \textbf{90\%}. FlexDoc offers a substantial scaling advantage: the annotation cost remains constant at ~150 minutes for 3,000 samples, while traditional hard-template methods require a linear increase in annotation time as the dataset size grows.

\subsection{Diversity Analysis}
\vspace{-8pt}
\begin{table}[ht]
\small
\renewcommand{\arraystretch}{1.3} 
\centering
\begin{tabular}{cc} 
  \hline
  \textbf{Dataset} & \textbf{Mean Pairwise Similarity} \\
  \hline
  DocILE - 5k Documents  & 0.52 ± 0.05 \\
  FlexDoc - 5k Documents & 0.55 ± 0.04 \\
  SROIE - 1k Documents    & 0.40 ± 0.02 \\
  FlexDoc - 1k Documents & 0.43 ± 0.04 \\
  \hline
\end{tabular}
\caption{\label{tab:pairwise-similarity}
  Mean pairwise similarity across different datasets, with standard deviation.
}
\end{table}

\vspace{-8pt}
We also evaluate the diversity of datasets generated by FlexDoc by comparing them to real-world benchmarks: DocILE and SRIOE. We compute mean pairwise similarity on datasets of comparable size (5000 samples for DocILE, 1000 for SRIOE), using embeddings from a LayoutLM model suited for document understanding. The results show that FlexDoc-generated data achieves diversity levels comparable to both real datasets. Moreover, the small deviations confirm the reliability of the results and further demonstrate that the generated documents are diverse without being dominated by outliers.

\section{Conclusion}
High-quality annotated data remains a major bottleneck for scaling document understanding in enterprise settings. We introduce a robust, extensible framework for generating realistic, diverse, and multilingual synthetic documents that mirror real-world complexity. FlexDoc enables rapid development of layout-aware models, greatly reducing the need for costly manual annotation. While our experiments target invoice-based KIE, the framework generalizes across document understanding tasks and is already in enterprise deployment, accelerating model development and significantly reducing data acquisition and annotation costs.

\section{Limitations and Future Work}
The data generator presented in this work aims to produce diverse documents by randomizing a large number of document attributes. While the permutations obtained by randomizing such a large number of variations work well for semi-structured documents, this approach is not particularly useful for fully structured, visually rich documents such as driving licenses, passports, etc. Such documents usually do not demonstrate much diversity, at least within a specific category. For example, driving licenses in the United States would have about 50-80 variations depending on the issuing state. The majority of the diversity is derived from variations in values (license holder's name, address, etc.) in such documents. These documents also contain visually rich backgrounds. For example, the driving license issued by Massachusetts contains an image of the State House in the background.

For these reasons, our approach is not suitable for generating synthetic documents for fully structured, visually rich documents. Hard-template-based approaches are still the most effective for such cases since the cost of annotation is not too high for documents with limited variations.

However, hard-template-based approaches have certain limitations as well, such as a loss of visual features from the background when filling in synthetic values. In the future, we would like to extend our approach to generate such documents while addressing the limitations of a hard-template-based approach. 

Additionally, FlexDoc doesn't generate semantically meaningful fake values. For example, the total amount in the invoice doesn't add up to amount of individual items. Therefore, we also plan to introduce semantic value generators in the future. Finally, we acknowledge that some type of documents may exhibit cultural variations and FlexDoc doesn't account for that when generating documents in other languages using the common stochastic schema.

\section*{Acknowledgments}
The work was conducted during employment with and funded by Oracle Corporation (AI Services).

\bibliography{custom}

\clearpage

\section*{Appendices}
\appendix
\label{sec:appendix}

\section{Experimentation Settings}
\label{sec:experimentation-settings}
\subsection{FlexDoc Generated Data}
For the purpose of experiments under \nameref{sec:evaluation}, the stochastic attributes in the JSON schema for FlexDoc generated data are defined using informed and reasonable approximations, however, the JSON can be modified to mimic other data distributions as well.

\subsection{Choice of Datasets and Models}
The experiments (for documents in English) are conducted using the DocILE dataset for benchmarking. DocILE is the largest open-source dataset in english language for key information extraction (KIE). Alternative datasets such as KVP10k and SROIE also exist, but they are less suitable for our use case.

KVP10k contains semi-structured documents, but it lacks explicit document type labels (e.g., receipts, invoices). Since our approach generates targeted documents, knowing the document type with certainty is essential, making KVP10k unsuitable. Moreover, although KVP10k does have bounding boxes, it does not provide text-level classes; instead, it defines key-value relationships between text pairs. For e.g. it defines that the header text for purchase order number \textbf{12HOUGH1} is \textbf{Purchase Order} without specifying the predefined class. Such header text may differ across documents and therefore classification with this information is not feasible.

Another option is the SROIE \cite{Huang_2019} dataset, which focuses on receipts. However, the state-of-the-art performance on this dataset - \textbf{97.8\%} accuracy with LayoutLM - leaves little room for further improvement through synthetic data augmentation.

We supplement our results with the IDSEM \cite{sanchez2022idsem} dataset to support our claim that FlexDoc can generate synthetic multilingual documents for training document understanding models.

We also conduct experiments using LayoutLM, a conventional encoder-based multimodal model, and Phi-4-Multimodal-Instruct, a modern decoder-based generative model, to evaluate the effectiveness of FlexDoc across both architectural paradigms and model types.

\subsection{Downstream Model Performance}
For training LayoutLM, the annotations in the dataset are chunked at word level using the IOB format \cite{ramshaw-marcus-1995-text}. Since the annotations are chunked at the word level using the IOB format, the evaluation is also conducted at the word level, without taking into account the I and B modifiers. For instance, in the case of a Merchant Name like \enquote{Jake Peralta},  both \enquote{Jake} and \enquote{Peralta} are treated as part of the \enquote{MerchantName} class, rather than as separate \enquote{B-MerchantName} and \enquote{I-MerchantName} tags.

For training and evaluating Phi-4-Multimodal-Instruct, the following prompt was used: \textit{Extract only the values for the following keys from the document image. If a key has multiple values, list all separated by a pipe (|). Return output in the following format as JSON: <json\_format>.}

\subsection{Ablation Study - Dynamic Virtual Grid Algorithm}
For this study, the Dynamic Virtual Grid algorithm is replaced with a simpler approach that places text in non-overlapping regions on a blank canvas. The canvas dimensions are kept consistent with those used in the overall FlexDoc algorithm. All other aspects of the data generation process remain unchanged, ensuring that the generated data still includes annotations such as bounding boxes and class labels.

\label{sec:appendix-hard-template}
\subsection{Comparison with Hard-Template Approach}
For the baseline, we generate 3000 samples using 300 manually annotated hard templates. We then generate the same number of samples using our own approach. To be fair to the baseline, we generate 10 copies of each schema frozen by our approach, effectively generating 3000 samples from 300 document permutations. We then train both models using these two datasets. The savings in annotation time is considered based on an average of 5 minutes spent per hard template (baseline) and 2.5 hours spent per stochastic schema (FlexDoc).

\subsection{Diversity Analysis}
To compute the diversity of datasets in this study, we utilize LayoutLM to extract joint representations that incorporate text, spatial layout (bounding boxes), and visual features. These representations are taken from the penultimate layer of the LayoutLM model. We then compute the mean pairwise cosine similarity between all document embeddings in the dataset. A lower mean similarity indicates greater diversity, as it reflects a broader spread in the representation space.

Let \( \mathcal{D} = \{x_1, x_2, \ldots, x_N\} \) be a dataset with \( N \) document samples. Each document \( x_i \) is encoded into a joint embedding \( \mathbf{h}_i \in \mathbb{R}^d \) using the penultimate layer of the LayoutLM model, capturing textual, spatial, and visual information.

The mean pairwise cosine similarity (MPCS) is computed as:

\[
\text{MPCS}(\mathcal{D}) = \frac{2}{N(N-1)} \sum_{i=1}^{N-1} \sum_{j=i+1}^{N} \cos(\mathbf{h}_i, \mathbf{h}_j)
\]

where cosine similarity is defined as:

\[
\cos(\mathbf{h}_i, \mathbf{h}_j) = \frac{\mathbf{h}_i \cdot \mathbf{h}_j}{\|\mathbf{h}_i\| \, \|\mathbf{h}_j\|}
\]

A lower MPCS implies higher diversity in the dataset.

\subsection{Training Hyperparameters}
For each experiment, each model is trained on 4 x A100 GPUs for 40 epochs, with a learning rate of 2e-5, weight decay of 0.1, and a batch size of 8 per GPU core (effective batch size of 32). Phi-4-Multimodal was finetuned using Low-Rank-Adaption (LoRA) \cite{hu2021loralowrankadaptationlarge} training.

\subsection{DocILE dataset split}
The DocILE dataset consists of 6680 real annotated business document images. There are 55 keys that indicate key information in these documents. The dataset is split into 5392, 500 and 1000 training, validation and test multi-page images respectively. However, since the test split is not openly available, and the authors claim that for the test set, documents in both training and validation sets are considered as seen during training, we utilized the validation dataset for testing. We also divided the document images page-wise and set aside validation data from the training set. This resulted in a final train-validation-test split of 5392-1347-633. For all experiments involving the DocILE dataset, the test split is used for evaluation.

\section{Stochastic Schema Definition and Description of Attributes}
\label{sec:appendix-template-definition}

Following is an example of an entity group definition.

\begin{lstlisting}[language=json, basicstyle=\ttfamily\small]
"entity_groups": [
    {
      "name": "DeliveryDetails",
      "segment": {
        "0": 0.3,
        "1": 0.3,
        "2": 0.3,
        "4": 0.03,
        "5": 0.03,
        "6": 0.03
      },
      "tabulate": {
        "create": 0.3,
        "rows": 1,
        "tabType": [
          "horizontal",
          "vertical"
        ]
      },
      "header": [
        "Delivery Information",
        "Delivery Info",
        "Delivery Details"
      ],
      "entities": [
        {
          "name": "customer_delivery_name",
          "align": [
            "left",
            "right",
            "center"
          ],
          "header_align": [
            "left",
            "right",
            "center"
          ],
          "probability": 1,
          "header": [
            "Recipient Name",
            "Recipient",
            "Customer Recipient",
            "C/O"
          ],
          "type": "company"
        },
        {
          "name": "customer_delivery_address",
          "probability": 0.5,
          "align": [
            "left",
            "right",
            "center"
          ],
          "header_align": [
            "left",
            "right",
            "center"
          ],
          "header": [
            "Delivery Address",
            "Customer Delivery Address",
            "Recipient Address",
            "Ship To",
            "Shipped To"
          ],
          "type": "address_multi_line"
        }
      ],
      "headerProbability": 1,
      "probability": 0.5
    }
]
\end{lstlisting}

Table~\ref{tab:entity-group-definition} provides a description of all the attributes for entity groups and Table~\ref{tab:common-config} for common configuration that can be defined in a stochastic schema JSON. JSON Definition for common configuration has been omitted for brevity. 

\onecolumn

\begin{longtable}{p{0.25\textwidth} p{0.7\textwidth}}
\caption{Description of attributes for Entity Group Definition} \\
\toprule
\textbf{JSON Key in Schema} & \textbf{Description} \\
\midrule
\endfirsthead

\toprule
\textbf{JSON Key in Schema} & \textbf{Description} \\
\midrule
\endhead

\bottomrule
\endfoot
\bottomrule
\endlastfoot

name & The name of the entity group. \\

segment & Probabilities for the entity group to appear in various page segments. The sum should be 1. The algorithm uses these probabilities to decide the location during runtime. \\

tabulate & Probability of creating the entity group as a table. Runtime logic uses this probability to decide layout. Table orientation (horizontal/vertical) is determined randomly based on the \texttt{tabType} key. The number of rows and empty rows are either defined by \texttt{rows} / \texttt{numEmptyRows}, or sampled randomly. \\

header & One header value from the list is selected randomly during runtime and written as the group header. \\

headerProbability & Probability of this entity group having a header. Used by the algorithm at runtime. \\

probability & Probability of this entity group being present in the document. Used at runtime to decide inclusion. \\

gridPosition & Overrides the Dynamic Virtual Grid Algorithm for placing this entity group at a fixed location (e.g., for Bill Details). Other groups are placed accordingly. \\

groupAlignment & List of alignment options. One is picked randomly, and applied to the group on the grid. \\

entities & 
\begin{itemize}
    \item \textbf{\texttt{name}}: Used as a label in the final annotation.
    \item \textbf{\texttt{probability}}: Probability that this entity is present, used at runtime.
    \item \textbf{\texttt{header}}: One header from the list is randomly selected and displayed.
    \item \textbf{\texttt{type}}: Type of entity. Guides fake value generation.
    \item \textbf{\texttt{fontVariance}}: Overrides for font face, size, and color. Unspecified properties remain unchanged.
    \item \textbf{\texttt{addHeader}}: Enforces header display for critical fields like "Total Amount" regardless of global choice.
    \item \textbf{\texttt{align}}: List of alignment options (applicable only when group is a table).
    \item \textbf{\texttt{headerAlign}}: Same as \texttt{align}, but for headers only (table layout only).
\end{itemize} \\

entityShuffleGroups & Defines subgroups of entities that can be shuffled among each other at runtime. Each subgroup is a list of entity names. \\
\label{tab:entity-group-definition}
\end{longtable}

\begin{longtable}{p{0.35\textwidth} p{0.6\textwidth}}
\caption{Description of attributes for Common Configuration Definition} \\
\toprule
\textbf{JSON Key in Schema} & \textbf{Description} \\
\midrule
\endfirsthead

\toprule
\textbf{JSON Key in Schema} & \textbf{Description} \\
\midrule
\endhead

\bottomrule
\endfoot
\bottomrule
\endlastfoot

faker\_locale & Locale for generating fake values using Faker. Enables multilingual support by setting the appropriate locale for the Faker instance. \\

translation & Enables machine translation of entity headers. Includes the following keys: \texttt{"enable"} (boolean as string) to toggle translation, and \texttt{"target\_lang\_code"} to specify the target language (e.g., "pt" for Portuguese). \\

fake\_value\_generator\_class & Fully qualified path to the custom fake value generator class. \\

structural\_config & 
\begin{itemize}
    \item \textbf{num\_segments}: Number of document segments.
    \item \textbf{segment\_size}: Rows and columns per segment.
    \item \textbf{canvas\_width}: Width of the blank canvas.
    \item \textbf{canvas\_height}: Height of the blank canvas.
    \item \textbf{intra\_group\_y\_offset}: Vertical spacing between entities in a group.
    \item \textbf{intra\_group\_x\_offset}: Horizontal spacing between entities in a group.
    \item \textbf{inter\_group\_y\_offset}: Vertical spacing between different entity groups.
    \item \textbf{space\_width\_weight}: Weight factor used to determine spatial offsets.
\end{itemize} \\

font\_colors & List of possible font colors for entities and group headers. A random color is selected and applied globally unless overridden. \\

font\_size & Minimum and maximum font size range for entities and headers. \\

font\_dir & Directory path containing fonts. Can be different for headers and entities. \\

canvas\_color\_options & List of canvas background colors. One color is randomly selected during document generation. \\

table\_config & Styling configuration shared across all table-type entity groups. \\

show\_entity\_headers\_probability & Global probability for showing entity headers. Applies uniformly unless locally overridden by the entity configuration. \\

consistent\_patterns\_for\_values & Defines constraints for generating consistent values. For example, when multiple values share a common format (e.g., currency), the same currency symbol is reused across instances. Example: \texttt{["\$","€","£"]}. \\

expected\_keys & List of all possible entities to include in annotations. Only those in this list are labeled with their specific names; others are marked as \enquote{Other}. \\
\label{tab:common-config}
\end{longtable}

\renewcommand{\arraystretch}{1.4} 

\begin{longtable}{p{0.25\textwidth} p{0.7\textwidth}}
\caption{Stochastic Attributes and Attribute Sampling Process} \\
\toprule
\textbf{Stochastic Attribute} & \textbf{Sampling Process} \\
\midrule
\endfirsthead

\toprule
\textbf{Stochastic Attribute} & \textbf{Sampling Process} \\
\midrule
\endhead

\bottomrule
\endfoot
\bottomrule
\endlastfoot

Choice of an entity group to be present in the document &
Generate a random number from a uniform distribution. If this number is less than the specified entity group probability, include the entity group in the document. \\

Placing an entity group in a section of the page &
Driven by the \enquote{Dynamic Virtual Grid Algorithm} described below. \\

Choice of creating an entity group as a table or a stack of entities &
Generate a random number from a uniform distribution. If this number is less than the tabulation probability as defined in the entity group definition, create the entity group as a table. Table orientation (horizontal or vertical) is selected randomly from the options specified in the template. \\

Number of rows for an entity group defined as a table &
If a number is specified, it is used directly. If set to \enquote{random}, a value is sampled from a uniform distribution within bounds defined in the common configuration. \\

Number of empty rows for an entity group defined as a table &
If a number is specified, it is used directly. If set to \enquote{random}, a value is sampled from a uniform distribution within bounds defined in the common configuration. \\

Choice of an entity group header to be present in the entity group &
Generate a random number from a uniform distribution. If this number is less than the specified entity group header probability, include the header in the group. \\

Choosing an entity group header &
Randomly pick one value from the list of headers defined in the entity group schema. \\

Choice of an entity being present in the entity group &
Generate a random number from a uniform distribution. If this number is less than the entity’s specified probability, include the entity in the group. \\

Global entity header choice &
Generate a random number from a uniform distribution. If it is less than the defined threshold, include headers globally. Critical entities may override this behavior and always include headers. \\

Choosing an entity header &
Randomly pick one value from the list of headers provided for the specific entity. \\

Choosing entity alignment &
Randomly select one alignment option from the list defined for the entity. \\

Shuffling Entities in Sub-Groups &
For each defined subgroup, identify the indices of member entities. Shuffle these indices and reassign them to the original entities in that subgroup. \\

Choice for Global Font Face and Color &
Font face and color for entity text and headers (group/entity) are chosen randomly from a predefined list of fonts available in the project. \\

Table Style Attributes &
Stylistic attributes like header font face/color, row font face/color, and separator styles are selected randomly for each table in the document. \\
\label{tab:attribute-sampling}
\end{longtable}

\begin{figure*}[h!]
  \includegraphics[width=1.0\linewidth]{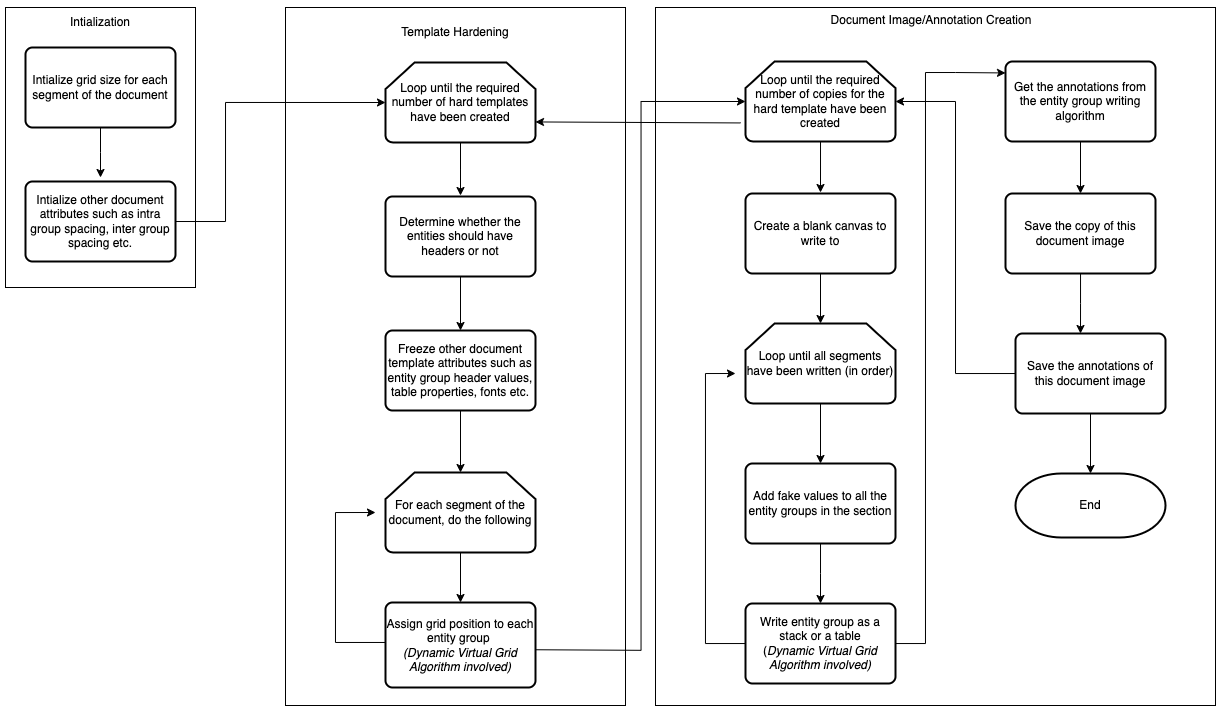}
  \caption {\label{fig:overall-algorithm}Overall Algorithm}
\end{figure*}

\begin{figure*}[h!]
  \includegraphics[width=1.0\linewidth]{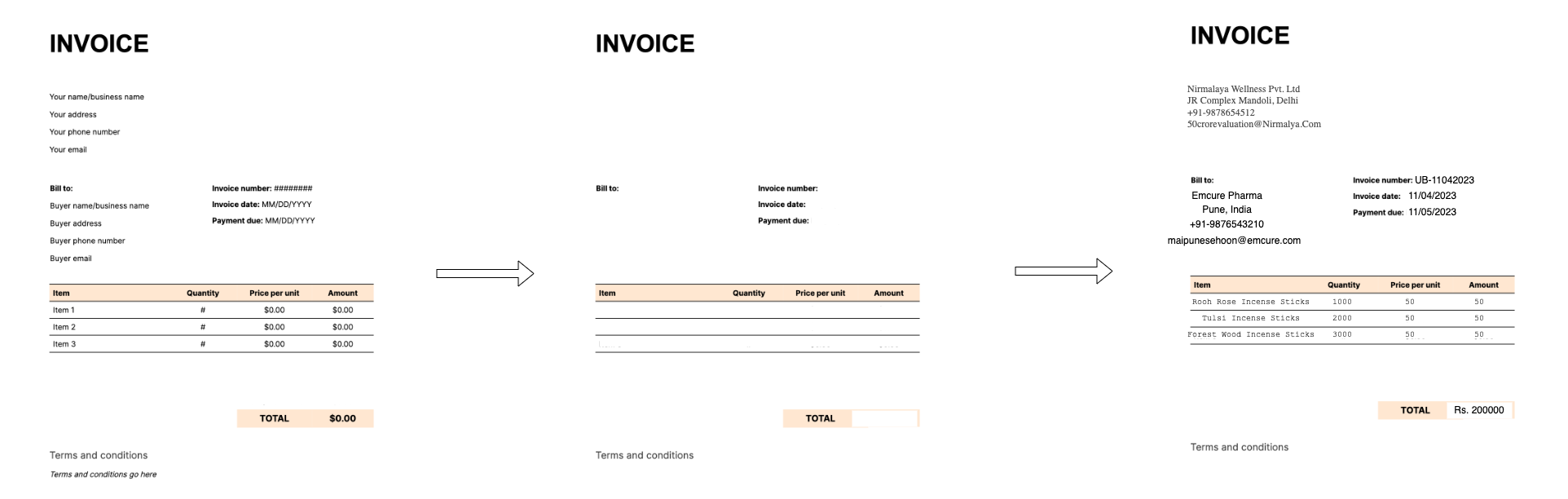}
  \caption {\label{fig:fully-templatic}Hard Template Based Synthetic Document Generation}
\end{figure*}

\begin{figure*}[h!]
  \includegraphics[width=1.0\linewidth]{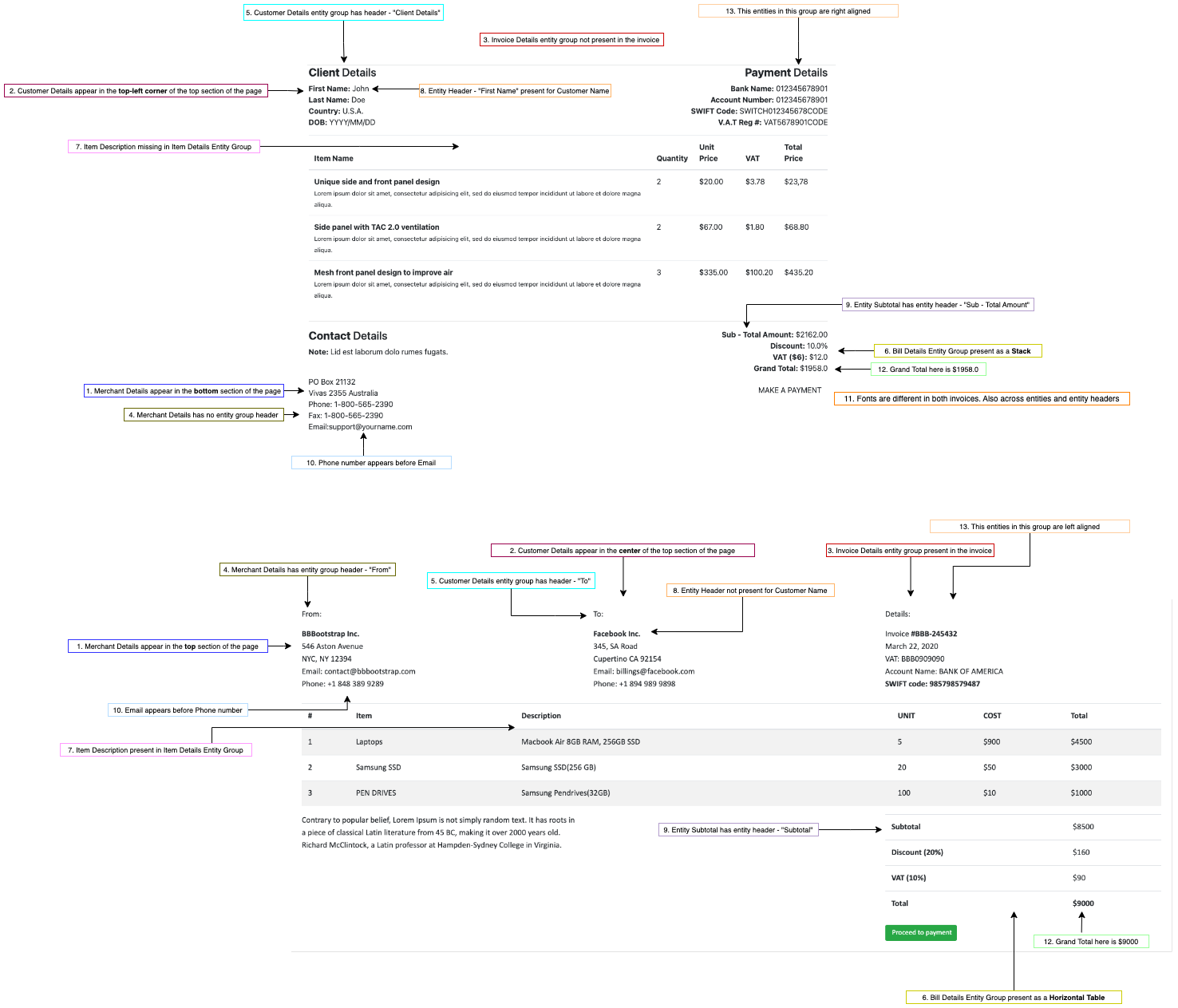}
  \caption {\label{fig:doc-variations}Two Invoices Depicting Variations in Semi Structured Documents}
\end{figure*}

\begin{figure*}[h!]
  \includegraphics[width=1.0\linewidth]{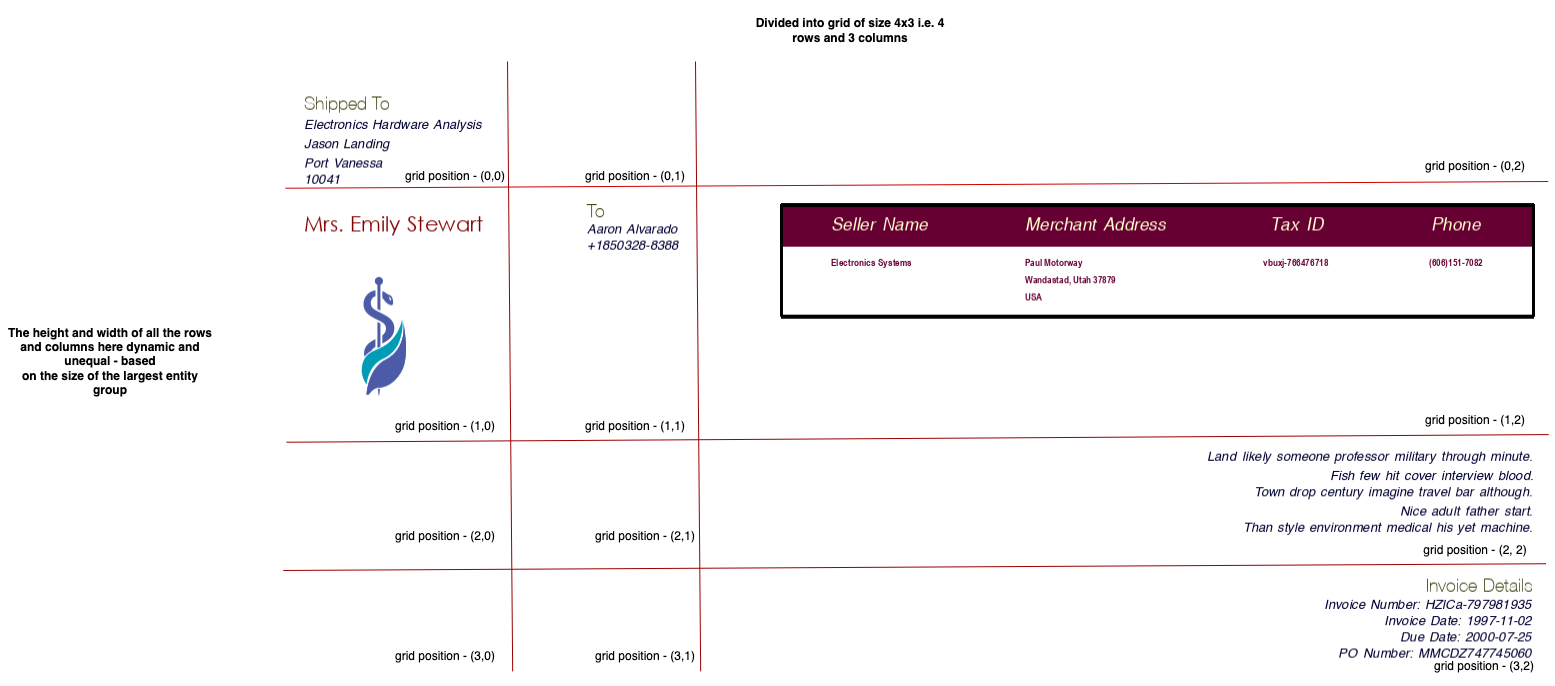}
  \caption {\label{fig:virtual-grid-algo-example}Arrangement of entity groups using Dynamic Virtual Grid Arrangement Algorithm. Do note that the arrangement is determined in-memory and the entities groups are written through a separate process.}
\end{figure*}

\begin{figure*}[h!]
  \includegraphics[width=1.0\linewidth]{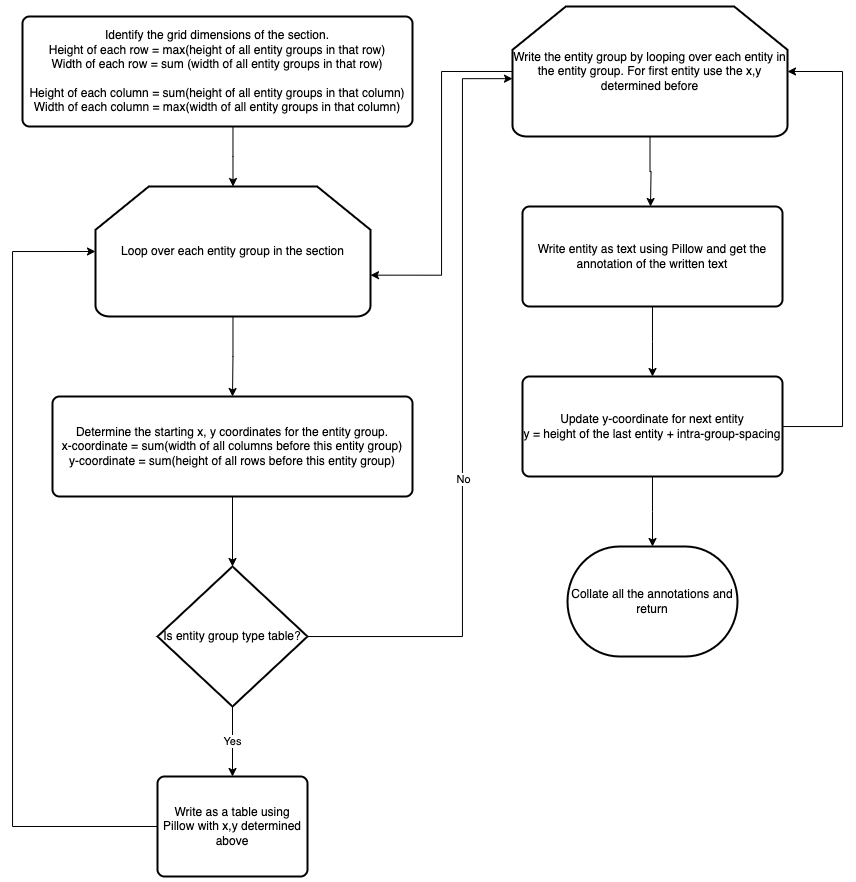}
  \caption {\label{fig:doc-rendering}Document Rendering Process}
\end{figure*}

\clearpage
\twocolumn
\section{Example Annotation JSON}
\label{sec:appendix-annotation}
Following is an example of two annotations generated by the framework. It contains a list of entities with their classes, along with tokenized child entities.
\begin{lstlisting}[language=json, basicstyle=\ttfamily\small]
[
    {
        "entity": [
            [
                132,
                38
            ],
            [
                377,
                27
            ],
            "SecureTrust Insurance"
        ],
        "children": [
            [
                [
                    132,
                    38
                ],
                [
                    206,
                    27
                ],
                "SecureTrust"
            ],
            [
                [
                    344,
                    38
                ],
                [
                    165,
                    27
                ],
                "Insurance"
            ]
        ],
        "class": "InsurerName"
    },
     {
        "entity": [
            [
                387,
                205
            ],
            [
                105,
                11
            ],
            "Scott Williams"
        ],
        "children": [
            [
                [
                    387,
                    205
                ],
                [
                    42,
                    11
                ],
                "Scott"
            ],
            [
                [
                    432,
                    205
                ],
                [
                    60,
                    11
                ],
                "Williams"
            ]
        ],
        "class": "MemberName"
    }
]
\end{lstlisting}
\onecolumn
\section{Generated Examples}
\label{sec:generated-samples}
\subsection{Invoices}

\begin{figure}[h!]
    \centering
    \begin{minipage}{0.35\textwidth}
        \centering
        \includegraphics[width=\textwidth]{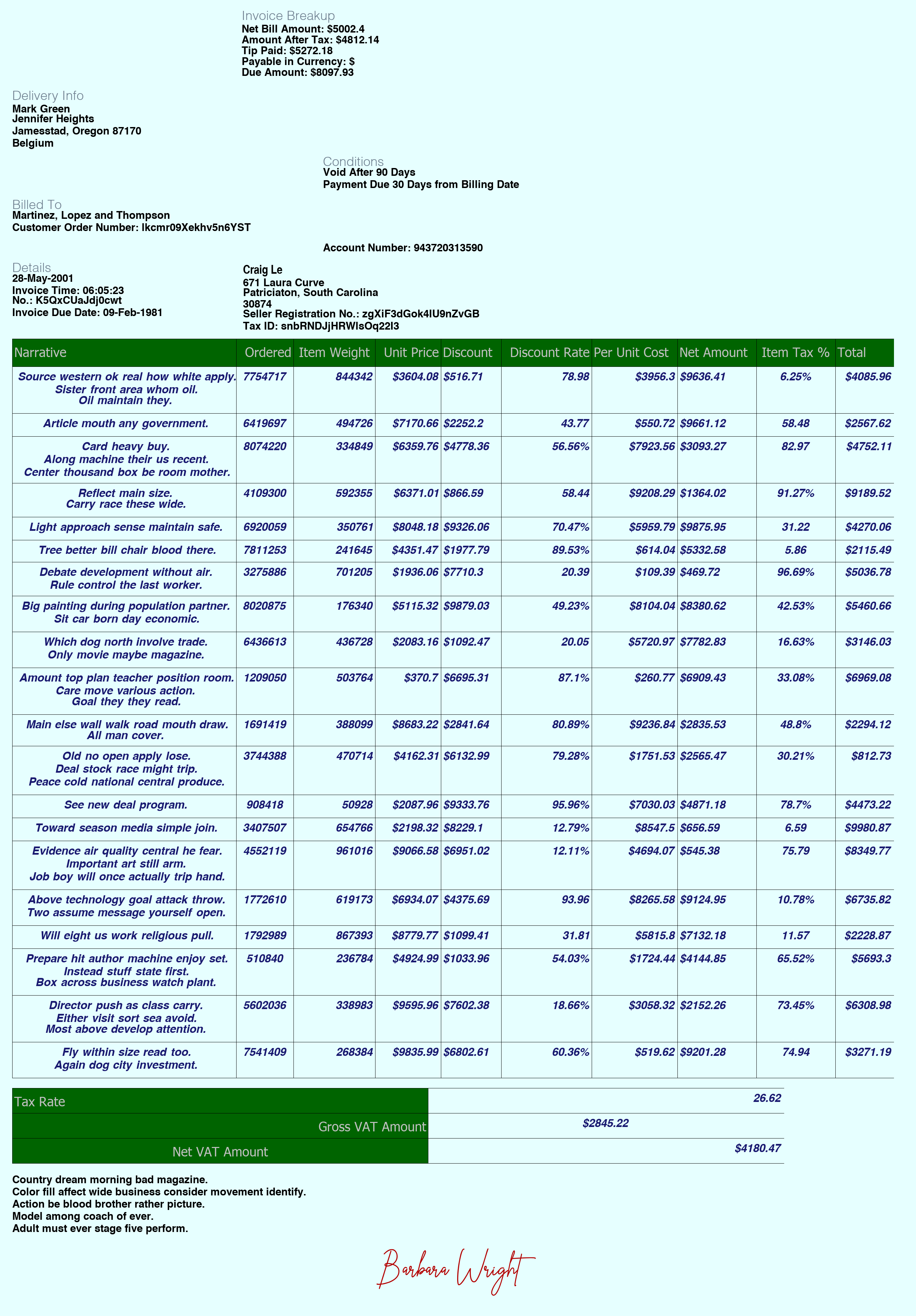}  
    \end{minipage}%
    \hfill
    \begin{minipage}{0.35\textwidth}
        \centering
        \includegraphics[width=\textwidth]{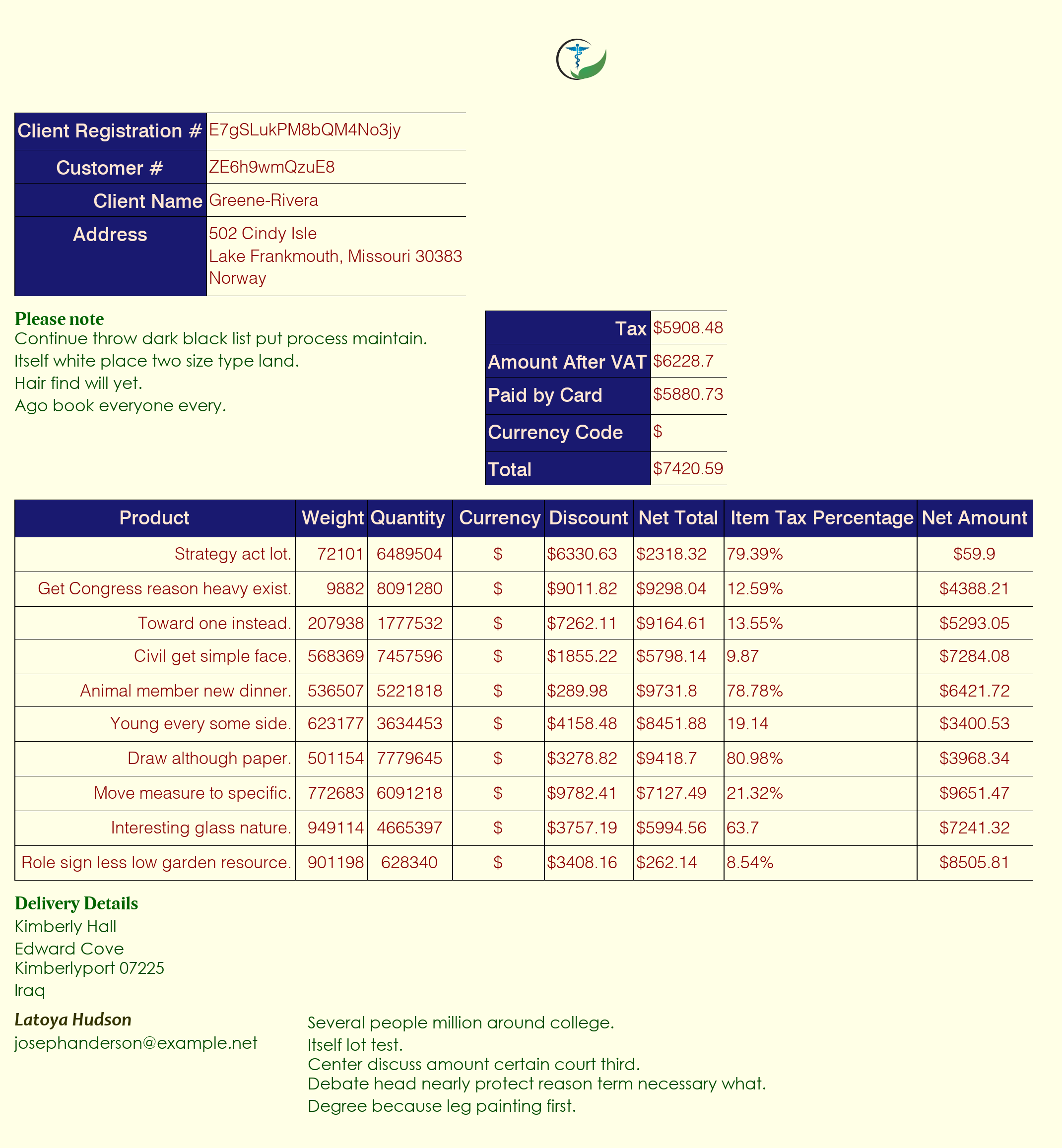}  
    \end{minipage}

    \vskip\baselineskip 

    \begin{minipage}{0.35\textwidth}
        \centering
        \includegraphics[width=\textwidth]{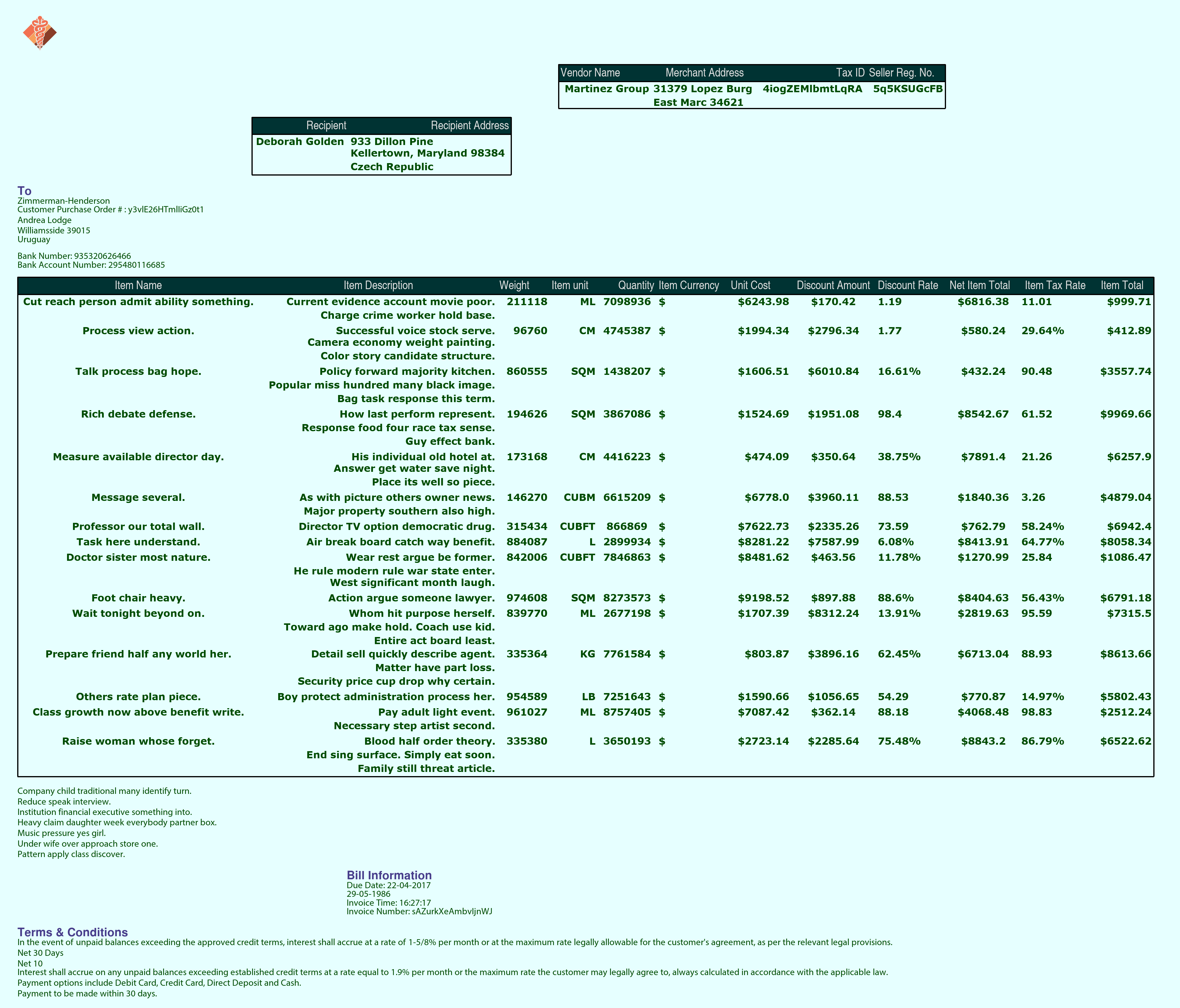}  
    \end{minipage}%
    \hfill
    \begin{minipage}{0.35\textwidth}
        \centering
        \includegraphics[width=\textwidth]{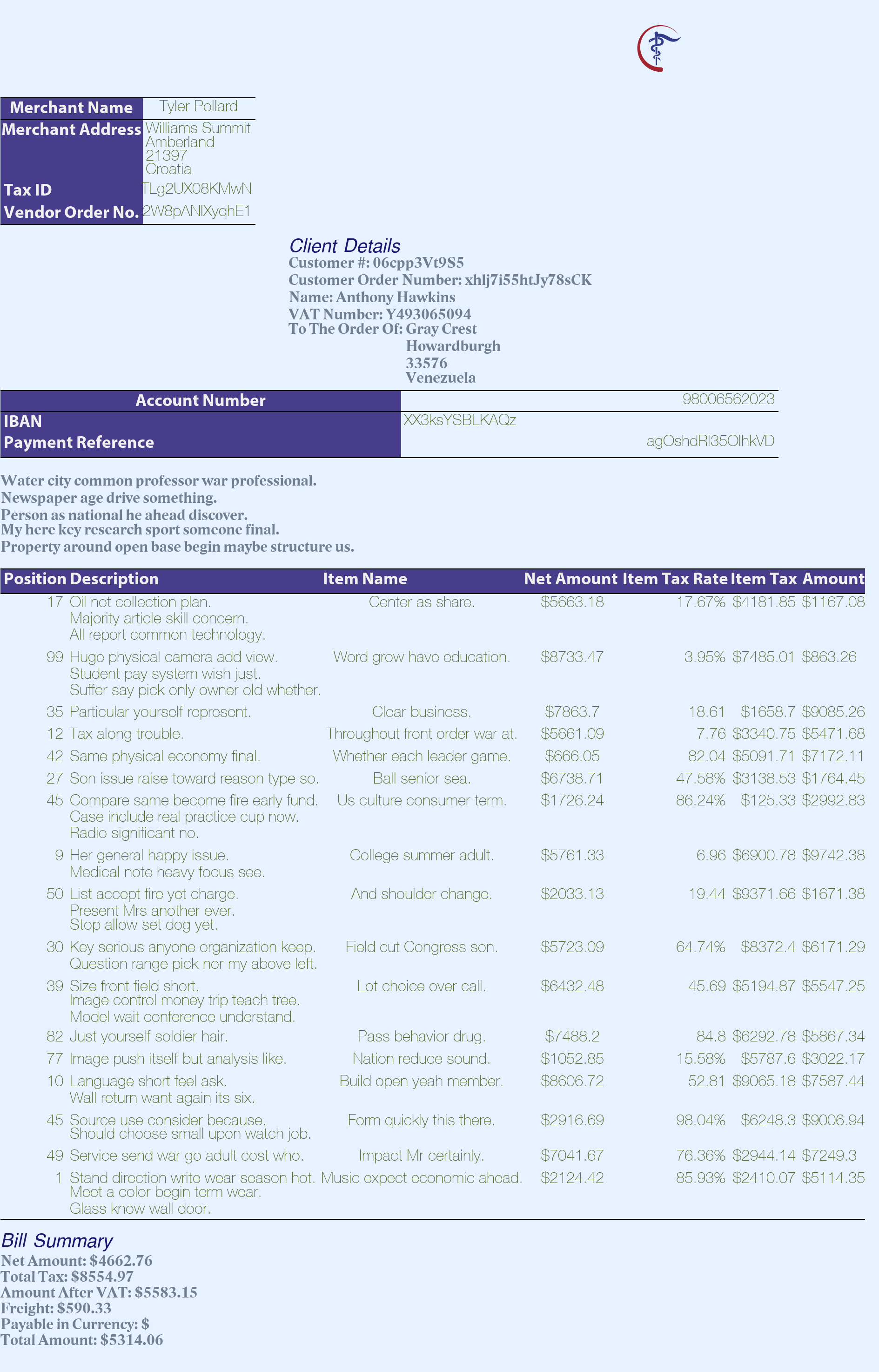}  
    \end{minipage}
    \caption{4 invoice images generated using the same Stochastic Schema. Notice the difference in values, structure of the images, the position and presence of entities and entity groups and overall styling and dimensions of the documents.}

\end{figure}

\pagebreak
\subsection{Insurance Cards}

\begin{figure}[h!]
    \centering
    \begin{minipage}{0.50\textwidth}
        \centering
        \includegraphics[width=\textwidth]{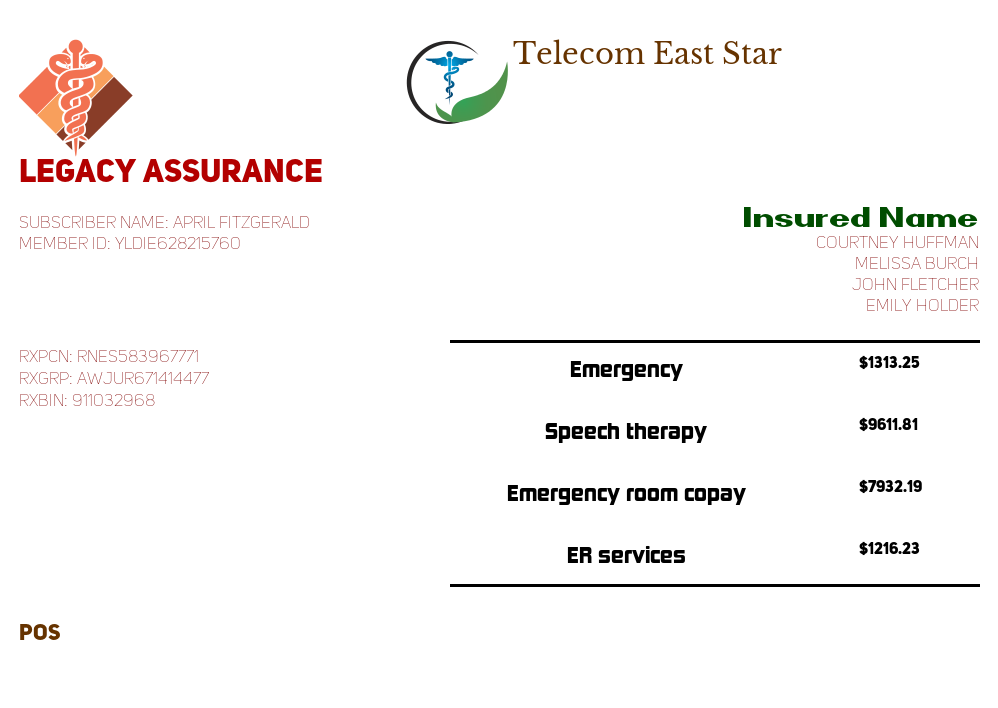}  
    \end{minipage}%
    \hfill
    \begin{minipage}{0.50\textwidth}
        \centering
        \includegraphics[width=\textwidth]{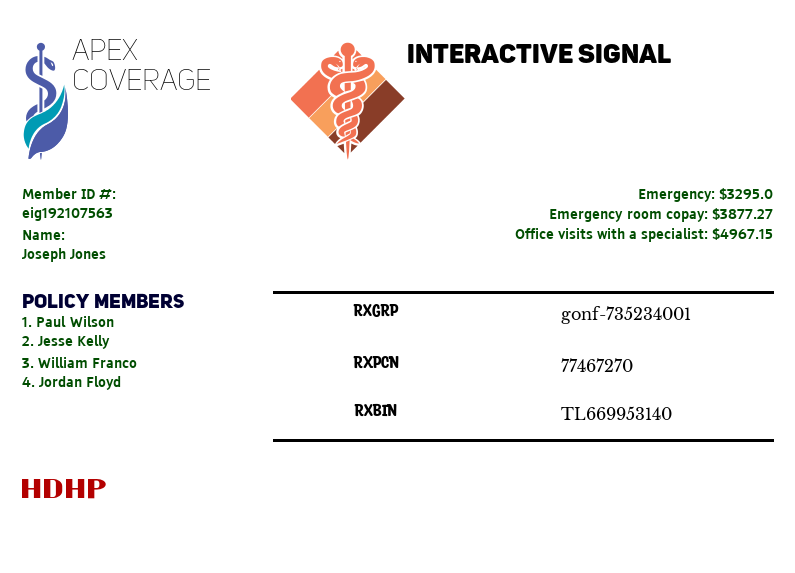}  
    \end{minipage}

    \vskip\baselineskip 

    \begin{minipage}{0.50\textwidth}
        \centering
        \includegraphics[width=\textwidth]{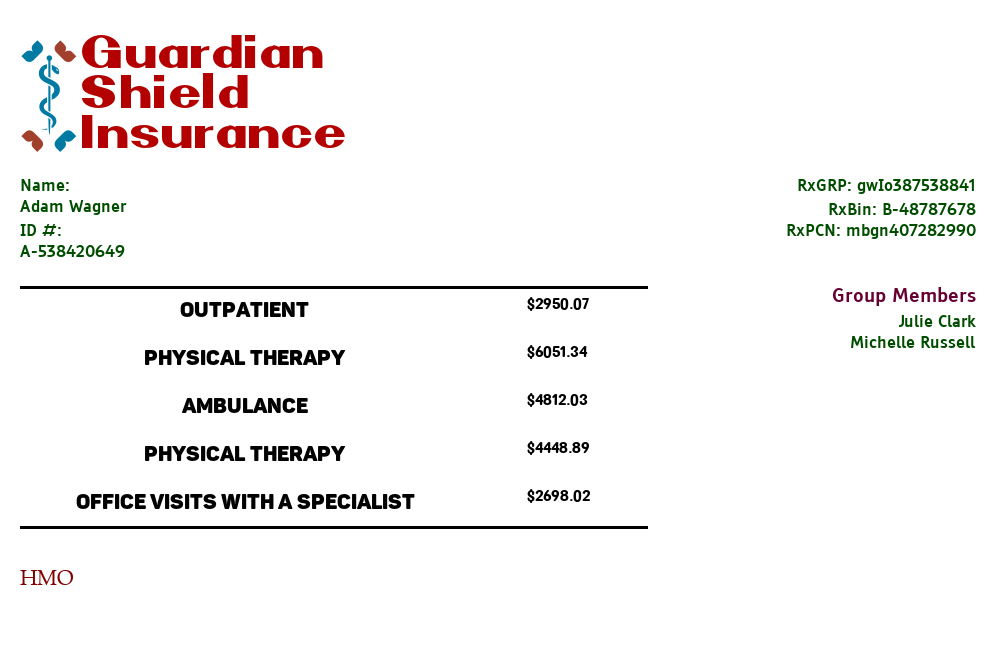}  
    \end{minipage}%
    \hfill
    \begin{minipage}{0.50\textwidth}
        \centering
        \includegraphics[width=\textwidth]{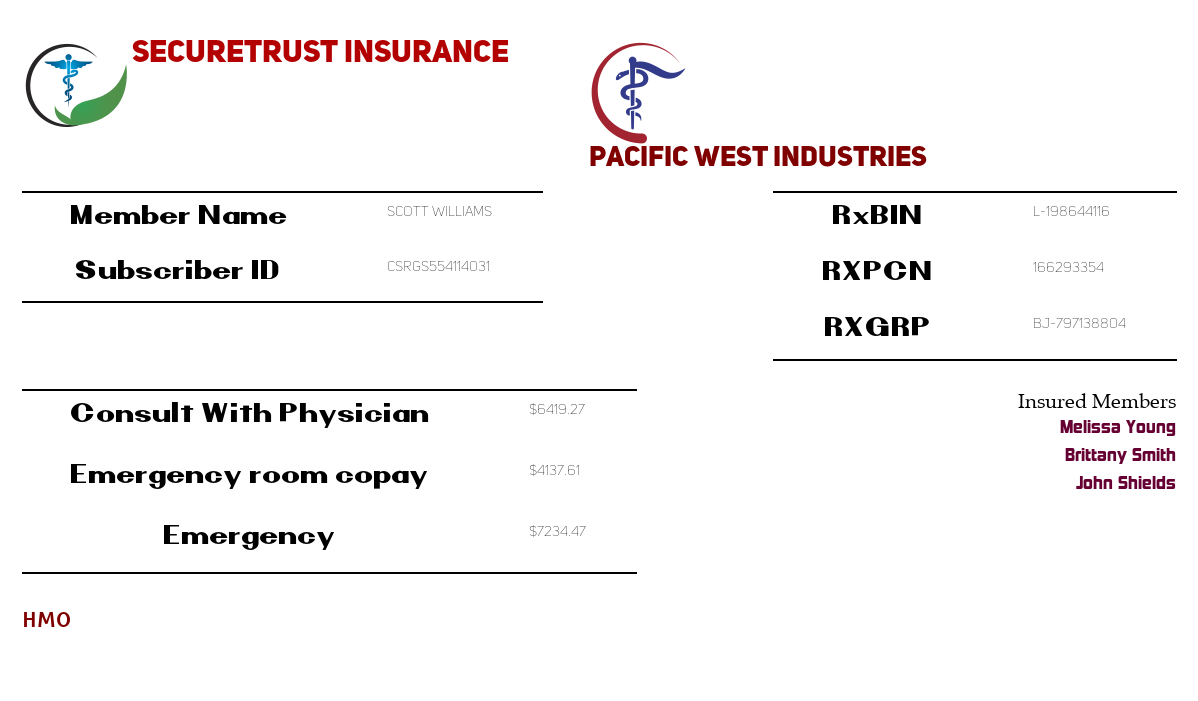}  
    \end{minipage}
    \caption{4 insurance card images generated using the same Stochastic Schema. Notice the difference in structure of the images, the position and presence of entities and entity groups and overall styling and dimensions of the documents.}

\end{figure}

\pagebreak

\end{document}